\documentclass[runningheads]{llncs}
\usepackage{graphicx}
\usepackage{amsmath,amssymb} 
\usepackage{subfigure}
\usepackage{caption}
\usepackage{booktabs}       
\usepackage{amsfonts}       
\usepackage{nicefrac}       
\usepackage{microtype}      
\usepackage{multirow}
\usepackage{color}
\usepackage[width=122mm,left=12mm,paperwidth=146mm,height=193mm,top=12mm,paperheight=217mm]{geometry}
\begin{document}
\pagestyle{headings}
\mainmatter
\def\ECCV18SubNumber{2659}  

\title{Pedestrian-Synthesis-GAN: Generating Pedestrian Data in Real Scene and Beyond} 

\titlerunning{}

\authorrunning{}

\author{
  Xi Ouyang$^{1,*}$ \hspace{0.7em} Yu Cheng$^{2,*}$ \hspace{0.7em} Yifan Jiang$^1$ \hspace{0.7em} Chun-Liang Li$^3$ \hspace{0.7em} Pan Zhou$^1$
}
\institute{$^1$ Huazhong University of Science and Technology,\hspace{1em}   $^2$IBM Research AI\\
$^3$Machine Learning Department, Carnegie Mellon University\\
  ($^*$ denotes equal contribution)}

\maketitle

\begin{abstract}
State-of-the-art pedestrian detection models have achieved great
success in many benchmarks. However, these models require lots of
annotation information and the labeling process usually takes much
time and efforts. In this paper, we propose a method to generate
labeled pedestrian data and adopt them to support the training of
pedestrian detectors. The proposed framework is built on the
Generative Adversarial Network (GAN) with multiple discriminators,
trying to synthesize realistic pedestrians and learn the background
context simultaneously.  To handle the pedestrians of different
sizes, we adopt the Spatial Pyramid Pooling (SPP) layer in the
discriminator. We conduct experiments on two benchmarks. The results
show that our framework can smoothly synthesize pedestrians on
background images of variations and different levels of details. To
quantitatively evaluate our approach, we add the generated samples
into training data of the baseline pedestrian detectors and show the
synthetic images are able to improve the detectors'
performance.\footnote{the code is available at
https://github.com/yueruchen/Pedestrian-Synthesis-GAN}
\end{abstract}

\section{Introduction}

Pedestrian detection is a crucial task in computer vision with a wide range of applications, including autopilot, surveillance and robotics \cite{enzweiler2009monocular,dollar2009pedestrian,dollar2012pedestrian,zhang2016far}. Recently, pedestrian detectors based on convolutional neural networks (CNNs), such as Faster R-CNN \cite{faster_rcnn} and YOLO9000 \cite{redmon2016yolo9000}, have been applied to various of benchmarks. Built on tremendous amount of training examples, these models can achieve significant performance improvement over previous baselines.

However, labeling ground-truth bounding boxes for pedestrian locations requires time consuming and  considerable human effort. Meanwhile, the performance of CNN-based pedestrian detectors heavily depends on the quality and the diversity of annotations in the training datasets. In other words, those methods expect the training data set to cover the same scenes or similar background environment as the testing data, such as camera configurations, lighting conditions and backgrounds. This becomes an issue when one applies these methods to a new unannotated video or video with limited supervision.
Therefore, it is very important to design approaches that only rely on limited supervision and can be extended to new unannotated datasets smoothly.

One way to solve this problem is to develop methods to automatically generate labeled datasets. There exists some efforts that use simulation techniques to generate pedestrian appearance and its location in the image \cite{cheung2016lcrowdv,hattori2015learning}. But these methods apply in strict environment like fixed cameras. One model proposed to be used for the moving camera \cite{cheung2017std}, can combine real-world background information in a scene with synthetically generated pedestrians. Nevertheless, since they generate pedestrians through rendering of 3D human models, the synthetic images look unrealistic and unnatural.

\begin{figure}[!t]
\centering
\includegraphics[width=12cm, height=3cm]{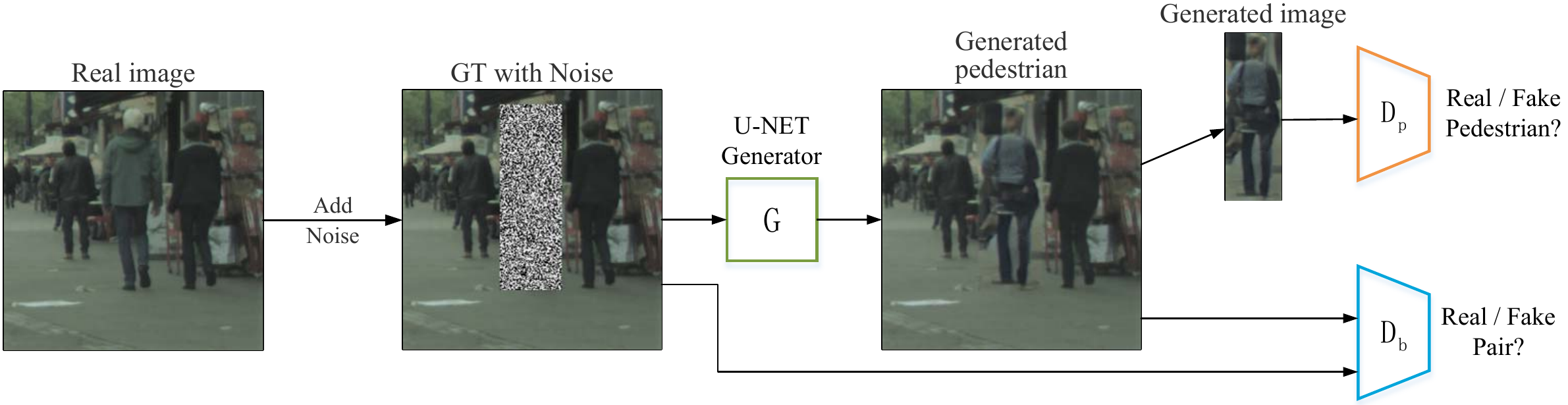}
\caption{The PS-GAN model learns to smoothly synthesize pedestrians in background images through the multiple discriminators ($D_b$ and $D_p$) network.}
\label{fig:framework}
\vspace{-.8em}
\end{figure}

Motivated by recent promising success of generative adversarial networks (GANs) \cite{goodfellow2014generative} in several applications \cite{isola2016image,pathak2016context,ledig2016photo}, we propose to build a GAN-based model to generate realistic pedestrian images in real scene and utilize them as the augmented data to train the CNN-based pedestrian detector. Compared with adopting the regular GAN as a powerful tool for generating images, the goal of our model is different and more challenging due to: 1) generating pedestrians to fit the background scene well; 2) providing the corresponding locations of those synthetic pedestrians as the ground truths for the CNN-based detectors. We denominate it as Pedestrian-Synthesis-GAN (PS-GAN).

PS-GAN adopts the adversarial learning recipe and contains multiple discriminators ($D_b$ for background context learning and $D_p$ for pedestrian classifying), as shown in Figure \ref{fig:framework}. We replace the pedestrians in the bounding boxes with random noise and train the generator $G$ to synthesize new pedestrians within that noise region. The discriminator $D_b$, learns to discriminate between real and synthesized pair. Meanwhile, the discriminator $D_p$ learns to judge whether the synthetic pedestrian cropped from the bounding boxes is real or fake. $D_b$ aims to force $G$ to learn the background information like the road, light condition in noise boxes. It leads to smooth connection between the background and the synthetic pedestrian. $D_p$ makes $G$ to generate real pedestrians with more realistic shape and details. Moreover, due to the varied sizes of cropped synthetic pedestrians, we utilize the Spatial Pyramid Pooling (SPP) layer \cite{he2014spatial} in $D_p$ to avoid the effect of resizing. After training, the generator $G$ can learn to generate photo-realistic pedestrians in the noise box regions and the locations of noise boxes are taken as the ground truths for detectors.

To the best of our knowledge, PS-GAN is the first work that utilizes GAN to generate data for pedestrian/object detection task. We evaluate it on two large-scale datasets: Cityscapes \cite{cityscapes} and Tsinghua-Daimler Cyclist Benchmark \cite{Tsinghua}. We use the model to generate results on these two datasets, and also train the Faster R-CNNs \cite{faster_rcnn} with real and synthetic data to prove the effectiveness of data augmentation. We show that:
\begin{itemize}
\item Our proposed model can generate sharp and photo-realistic pedestrian images and fit the background well in real scene/image;
\item The data generated from PS-GAN can be used with some real samples to train CNN-based detectors. This data augmentation step can improve both detection performance and stability over original model;
\item On cross-dataset experiments, i.e., model is trained on one dataset and tested on the other, PS-GAN is also able to generate good samples and improve the performances of CNN-based detectors.
\end{itemize}

\section{Related Work}
\subsubsection{Pedestrian Detection} Pedestrian detection attracts great interest due to its wild applications including driving systems, surveillance and robotics \cite{enzweiler2009monocular,dollar2009pedestrian,dollar2012pedestrian,zhang2016far}.
Built upon parameterized CNN models, recent works \cite{faster_rcnn,redmon2016yolo9000,ouyang2013joint,cai2016unified,zhang2016faster} can achieve good detection performances in several benchmarks.
However, these models require a large amount of training samples, which is quite time-consuming and takes many human efforts.

To handle this issue, researchers have proposed different solutions, one of which is to develop data augmentation techniques. Existing data augmentation methods are generally limited to certain tasks or conditions: \cite{cheung2016lcrowdv} focuses exclusively on crowd behavior, \cite{hattori2015learning} works only when the camera is stationary. The paper \cite{cheung2017std} provides an automatic and relatively robust model STD-PD, which selects possible locations to place synthetic agents. Using a 3D model for pedestrian rendering, its generated pictures are not realistic. Realizing that it is difficult to model the complex distribution of  pedestrians in real scene by using hand-crafted rules only, we decide to adopt data-driven approach like GANs to perform the task.

\subsubsection{Generative Adversarial Network} The original GAN was proposed by \cite{goodfellow2014generative}, and there are plenty of following works to improve the training stability and visual quality of the generation \cite{zhaicfz16,radford2015unsupervised,mao2017lsgan,denton2015deep,arjovsky2017wasserstein,li2017mmd,gulrajani2017improved}. GANs also have been employed in many other applications, for example, super-resolution \cite{ledig2016photo}, image in-painting \cite{pathak2016context,yeh2017semantic,denton2016semi}, image translation \cite{isola2016image,taigman2016unsupervised,liu2017unsupervised,zhu2017unpaired}. \cite{radford2015unsupervised} proposed DCGAN and adopted it to augment training data for person re-identification, which focuses on verifying the effectiveness of the label smoothing regularization, instead of the quality of the generated pictures. \cite{ma2017pose} proposed PGGAN to synthesize persons in arbitrary poses in the cropped person images.


The most related work to ours is the $Pix2Pix$ GAN \cite{isola2016image},
which have solid and robust results when paired training samples are available. \cite{zhu2017unpaired} add Cycle Consistency Loss to the original version, enabling the model to conduct translations without paired training examples and task-specific designed functions. To synthesize the pedestrians in the noise boxes and the locations of which can be taken as the bounding box labels, we adopt the paired training in $Pix2Pix$ GAN but a different architecture with multi-discriminators.

Compared with the image in-painting work \cite{pathak2016context,yeh2017semantic,denton2016semi}, which aims to fill the randomly removed monochromatic patches in original image, our framework fills the missing area with noise rather than monochromatic blocks to generate patches with diverse shapes/colors. We only need to learn the background information based on the context provided by surrounding parts of the image when synthesizing pedestrians in noise boxes. The work in \cite{denton2016semi} exploits a similar two discriminators GAN for image in-painting to learn more context information of surrounding pixel. Different from that, we pass the pedestrian patch cropped from the generated output into the discriminator to encourage the model to generate person in diverse shapes.


\section{Pedestrian-Synthesis-GAN}
Generative Adversarial Networks \cite{goodfellow2014generative} consist of a generator $G$ and a discriminator $D$ that compete in a two-player minimax game. In this paper, we adopt the adversarial learning idea and propose PS-GAN with multiple discriminators, which has ability to synthesize photo-realistic pedestrians with the corresponding bounding boxes information. Unlike the regular GAN, our method leverages an adversarial process between the generator $G$ and both two discriminators: $D_b$ for background context learning and $D_p$ for discriminating pedestrian.

Our framework is inspired by the conditional GAN work \cite{mirza2014conditional}. While training, we replace the pedestrian region in original image with random noise and push it to the generator $G$. Suppose the noise image is $x$ while the original image with pedestrian is $y$. $G$ is trying to generate fake image from $x$ as similar as possible to $y$ to fool the two discriminators $D_b$ and $D_p$.
Therefore, when generating new data, we can place noise boxes on the certain area where the pedestrians are expected and use the generator $G$ to synthesize pedestrian within the noise boxes. In this section, we first introduce the model architecture, then the detailed formulation of the overall objective.

\begin{figure}[!t]
\centering
\includegraphics[width=12cm, height=5.5cm]{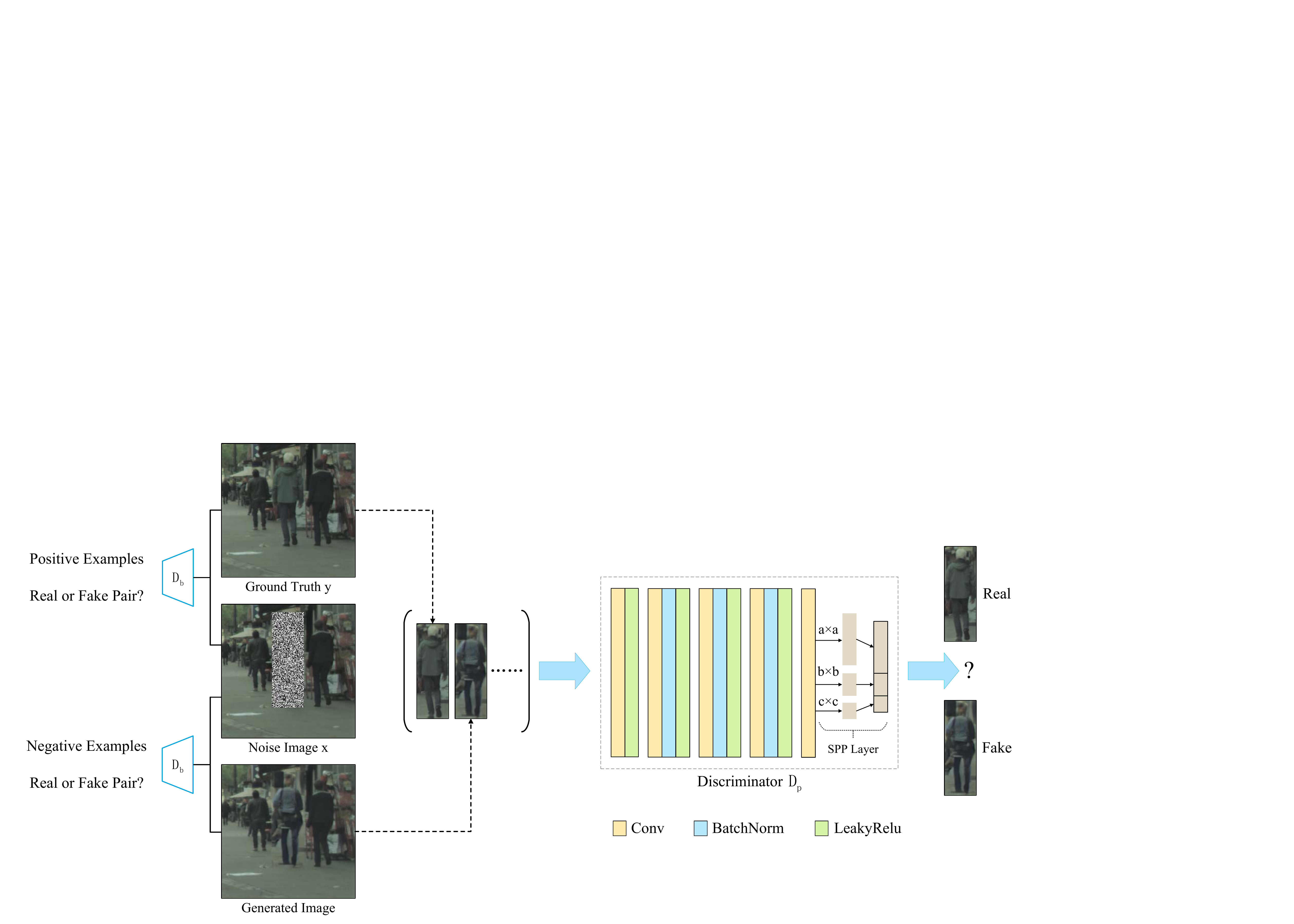}
\caption{The discriminator $D_b$ is applied to classify between real and synthesized pair to learn the background context in the noise box. The discriminator $D_p$ learns to classify the real and synthesized pedestrian with the noise box. We adopt a 3-level SPP layer ($a=1$, $b=2$, $c=4$, totally 21 bins) before the final feature representation.}
\label{fig:D}
\vspace{-.8em}
\end{figure}

\subsection{Model architecture}


\subsubsection{U-Net for Generator $G$}
The generator $G$ learns a mapping function $G: x \rightarrow y$, where $x$ is the input noisy image and $y$ is the ground truth image.
In this work, we adopt the enhanced encoder-decoder network (U-Net) \cite{isola2016image} for $G$. It follows the main structure of the encode-decoder architecture, where the input image $x$ is passed through a series of convolutional layers as downsampling layers until a bottleneck layer. Then the bottleneck layer feeds the encoded information of original inputs to the deconvolutional layers to be upsampled. U-Net uses the skip connections to connect the downsampling and upsampling layers in a symmetric position with respect to the bottleneck layer, which can preserve richer local information.


\subsubsection{$D_p$ to Discriminate fake/real Pedestrians}
For this discriminator $D_p$, we crop the synthetic pedestrian from the generated image as the negative example, while the real pedestrian $y_p$ from the original image $y$ as the positive example. Therefore, $D_p$ is used to classify whether the pedestrian is real or fake in the noise box. It forces $G$ to learn the mapping from the noise $z$ to the real pedestrian $y_p$, wher $z$ is the noise region in the noise image $x$.

The overall structure of $D_p$ is shown in Figure \ref{fig:D}. We apply a 5-layer convolutional network with LeakyRelu and BatchNorm layers. Normally the discriminator net accepts a fixed-size input. However, the input for our $D_p$ is the cropped pedestrian from the generated image or the ground truth image, which have various sizes. To address this issue, we adopt the Spatial Pyramid Pooling (SPP) layer \cite{he2014spatial} in $D_p$ and the detail of SPP-layer is also shown in Figure \ref{fig:D}. In our experiments, for each cropped pedestrian, we use a 3-level spatial pyramid ($1 \times 1$, $2 \times 2$, $4 \times 4$, totally 21 bins) to pool the features. After that, we concatenate all those 3-level features to an entire feature vector and apply the $Patch$ GAN loss \cite{isola2016image} here.

\subsubsection{$D_b$ to Learn Background Context}
The goal of our model is to not only synthesize a realistic pedestrian but also smoothly fill the synthetic pedestrian into the background. Thus it requires our model to learn context information like light conditions, surrounding backgrounds, etc. Following the pair-training recipe from $Pix2Pix$ GAN \cite{isola2016image}, $D_b$ is used to classify between real and synthesized pairs. The real pair concatenates the noise image $x$ and ground truth image $y$ while the synthesized pair concatenates the noise image $x$ and the generated image. The overall framework training $D_b$ is shown in Figure \ref{fig:D}.

The main structure of $D_b$ follows the design of DCGAN \cite{radford2015unsupervised} with the following modifications: 1) we make the first convolutional layer accept the 6-channel input of the stacked pair of images; 2) we use the $Patch$GAN in this discriminator as in \cite{isola2016image}, which means $D_b$ tries to classify if each $N \times N$ (in our experiment, $N$ is set to 70) patch in an image is real or fake; 3) we adopt the loss function of LSGAN \cite{mao2017lsgan} in $D_b$. To fit the $Patch$GAN setting, we calculate the mean squares between the $N \times N$ output and corresponding all-ones or all-zeros matrix as the loss function for $D_b$.

\subsection{Loss Function}
As illustrated in Figure \ref{fig:framework}, our model consists of two adversarial learning procedures $G \Leftrightarrow D_b$ and $G \Leftrightarrow D_p$. The adversarial learning between $G$ and $D_b$ can be formulated as:
\begin{multline}
\mathcal{L}_{LSGAN}(G,{D_b}) = {E_{y\sim{p_{gt \cdot image}}(y)}}[{({D_b}(y) - 1)^2}]
\\
+ {E_{x,z\sim{p_{noise \cdot image}}(x,z)}}[{({D_b}(G(x,z)))^2}],
\end{multline}
where $x$ is the image with the noise box and $y$ is the ground truth image. We use LSGAN here to replace the original GAN loss by a least square loss.

To encourage $G$ to generate realistic pedestrians within the noise box $z$ in the input image $x$, we conduce the other adversarial procedure between $G$ and $D_p$:
\begin{multline}
{\mathcal{L}_{GAN}}(G,{D_p}) = {E_{{y_p}\sim{p_{pedestrian}}({y_p})}}[\log {D_p}({y_p})]
\\
+ {E_{z\sim{p_{noise}}(z)}}[\log (1 - {D_p}(G(z)))],
\end{multline}
where $z$ is the noise box in $x$ and $y_p$ is the cropped pedestrian in the ground truth image $y$. We use the negative log likelihood objective to update the parameters of $G$ and $D_p$.

The training of GAN can be benefited from the traditional loss \cite{isola2016image}. In this paper, we apply $\ell _1$ loss to control the differences between the generated image and ground image $y$:
\begin{small}
\begin{equation}
{\mathcal{L}_{{\ell _1}}}(G) = {E_{x,z\sim{p_{noise \cdot image}}(x,z),y\sim{p_{gt \cdot image}}(y)}}[{\left\| {{\rm{y}}
- G(x,z)} \right\|_1}],
\end{equation}
\end{small}
Finally, combining the losses previously defined results in the final loss function as:
\begin{small}
\begin{equation}
\mathcal{L}(G, D_b, D_p) = \mathcal{L}_{LSGAN}(G,{D_b}) + {\mathcal{L}_{GAN}}(G,{D_p})+ \lambda {\mathcal{L}_{{\ell _1}}}(G).
\end{equation}
\end{small}
where $\lambda$ controls the relative importance of the $\ell _1$ loss. Empirically, we found that $\lambda = 100$ is a good setting and fixed it for the all experiments.

\begin{figure}[!t]
\centering
\includegraphics[width=12cm, height=5cm]{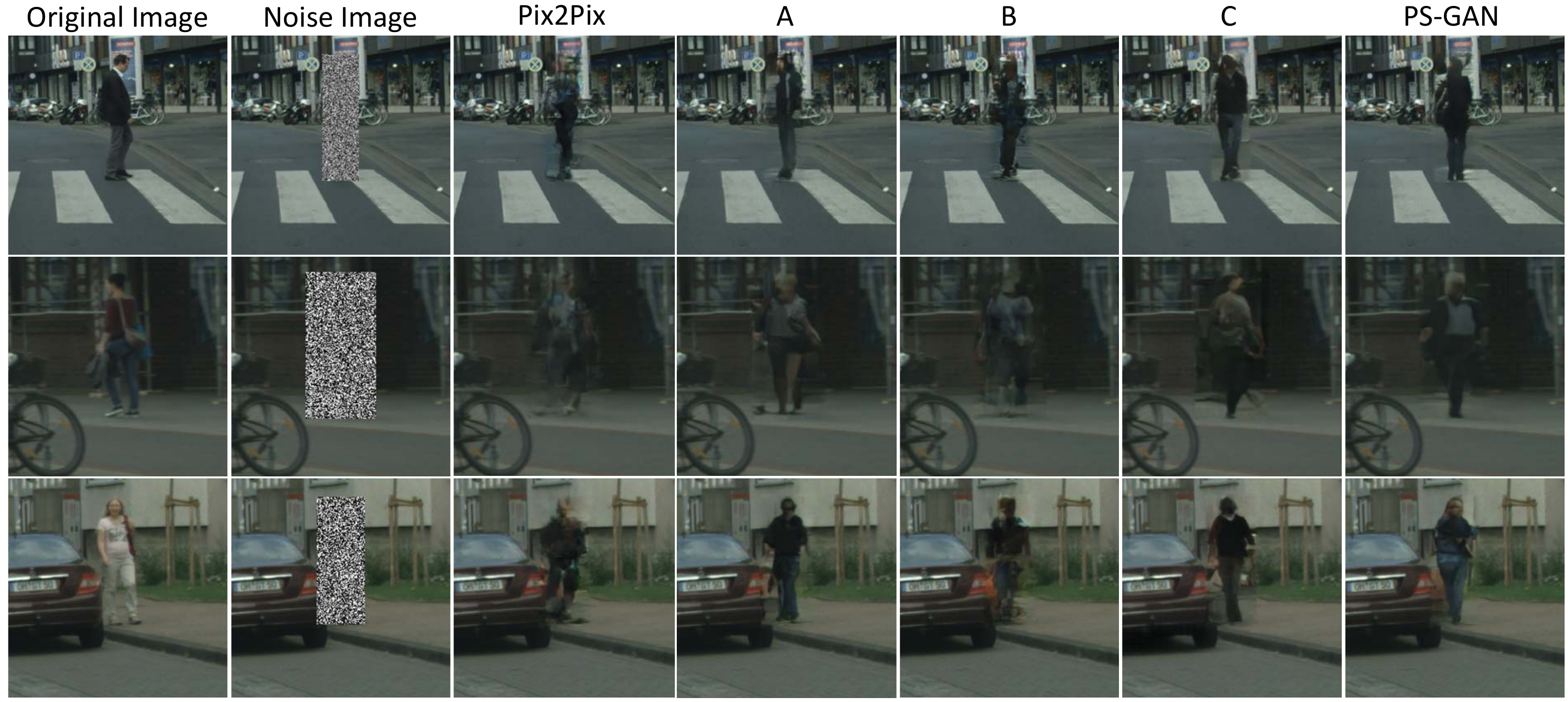}
\caption{We compare PS-GAN with four different models. The baseline is $Pix2Pix$ GAN \cite{isola2016image} in the third column, which only contain one discriminator to classify the real and synthesized pair. The following columns (A-C) show the ablation test of the proposed PS-GAN. Model $A$: the main structure is same as  PS-GAN but the SPP layer is removed; Model $B$: The difference with our final model is that this model adopt the LSGAN loss on both the two adversarial learning $G \Leftrightarrow D_b$ and $G \Leftrightarrow D_p$; Model $C$: The regular GAN loss is kept in both two adversarial learning procedures.
}
\label{fig:examples}
\vspace{-.8em}
\end{figure}

\section{Experimental Results}
We test PS-GAN model on Cityscapes \cite{cityscapes} and show the quality of  the synthesized images. To analyze the effect of the data augmentation, we combine the real and synthesized data to train the Faster R-CNN \cite{faster_rcnn} detectors and evaluate the performance. Moreover, to evaluate the ability to generate training example on the new video with limited supervision, we test PS-GAN model trained using Cityscapes on Tsinghua-Daimler Cyclist Benchmark \cite{Tsinghua}. All those experiments are based on PyTorch \footnote{\tt http://pytorch.org/} and run on Titan X GPUs.

\subsection{Cityscapes}
The Cityscapes dataset is a large-scale dataset for semantic urban scene understanding that contains a diverse set of stereo video recordings from 50 cities \cite{cityscapes}. Compared to other benchmarks like Caltech Pedestrian\cite{caltech} and KITTI\cite{Geiger2012CVPR}, Cityscapes has higher resolution pictures and contains more pedestrians with rich variety, which is more suitable to train GANs.

\subsubsection{Qualitative Result}
We generate the bounding boxes for all pedestrians based on the pixel-wise labels. There are some labeled pedestrians which are too small or partially blocked by cars or walls. So we filter out all the bounding boxes with the height smaller than 70 pixels and width smaller than 25 pixels. After that, we obtain 2326 images containing totally 9708 labeled pedestrians and randomly select 500 images of them as the testing dataset. We do not feed the original images ($1024 \times 2048$) into PS-GAN directly. Instead, we crop the $256 \times 256$ patches around the chosen pedestrians from the original $1024 \times 2048$ images. Moreover, we select 1200 pedestrian patches from the 1826 training images which display intact body shapes. Those 1200 patches will be covered with noise boxes in the pedestrian positions, then those noise images are taken as the training data for PS-GAN.

\begin{figure}[!t]
\centering
\includegraphics[width=12cm, height=5cm]{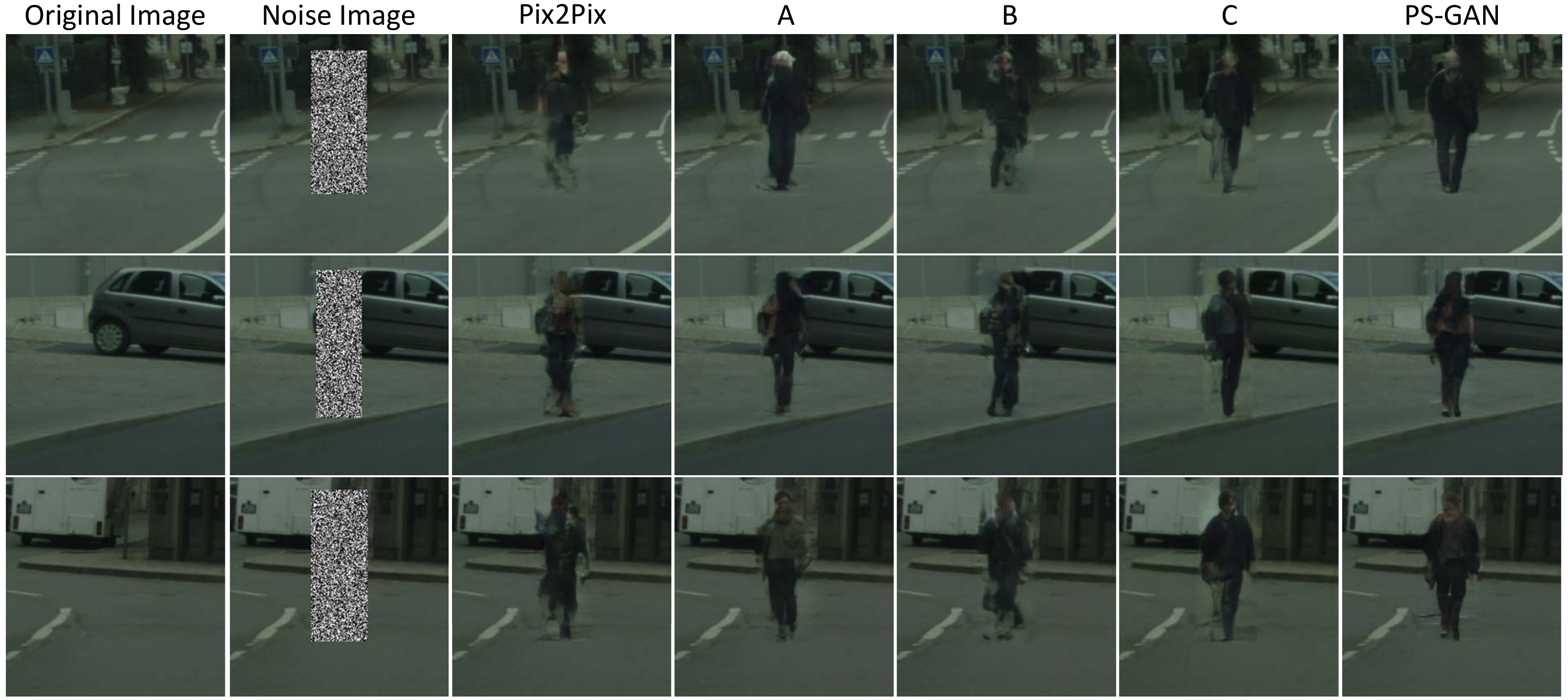}
\caption{Results of different models for synthesizing pedestrians in blank background.}
\label{fig:examples2}
\vspace{-.8em}
\end{figure}

To show the pedestrians generated by PS-GAN, we conduct two experiments: 1) generating pedestrians on the real pedestrian positions, and 2) generating pedestrians only on the background images without pedestrians. For the first setting, we crop the $256 \times 256$ patches around the pedestrians from the original $1024 \times 2048$ images among the 500 test examples and fill the noise boxes to cover the real pedestrians in those patches. Our pre-trained generator synthesizes pedestrians within those noise boxes and we compare the synthetic and real pedestrians as shown in Figure \ref{fig:examples}. For the second setting, we randomly crop the $256 \times 256$ patches from the blank scene images without any labeled pedestrians. Considering that the pedestrians can not appear in unreasonable positions like in the wall or within a car, we remove those wrong images and add the noise boxes in the remaining image patches. The results are in Figure \ref{fig:examples2}.

We list the synthesized samples of all baseline models, trained on the same training set for 200 epochs. Compared with the baseline $Pix2Pix$ GAN, PS-GAN can generate better quality of images both on Figure \ref{fig:examples} and Figure \ref{fig:examples2}. Most of the results of $Pix2Pix$ GAN only have murky person shapes while PS-GAN gets very clear shape of pedestrians. It proves that our discriminator $D_p$ can effectively guide generator $G$ to learn more realistic shape information and details of pedestrians. To evaluate the effect of the SPP layer in $D_p$, we compare the results of PS-GAN with the model $A$, which does not have SPP layer in $D_p$. As shown in Figure \ref{fig:examples} and Figure \ref{fig:examples2}, the model with SPP layer can learn more detailed information of pedestrians. For instance, in the first row of both Figure \ref{fig:examples} and Figure \ref{fig:examples2}, the legs of the person from PS-GAN can clearly be seen while they are blurry in the one from model $A$.

In our experiments, we find that using LSGAN \cite{mao2017lsgan} for $D_b$ is helpful to learn the background context. PS-GAN can obtain the best picture quality when applying the least square loss for the adversarial learning $G \Leftrightarrow D_b$ and keeping the regular GAN loss for $G \Leftrightarrow D_p$.
We design the model $B$ that adopts the LSGAN loss in both adversarial learning procedures, but the results are not competitive with PS-GAN as shown in both Figure \ref{fig:examples} and Figure \ref{fig:examples2}. Model $B$ performs only slightly better than the $Pix2Pix$ GAN.
We also study model $C$, which uses the regular loss on both adversarial learning procedures. Actually, the model $C$ can generate pedestrians with nice human-body shape. In the last row of Figure \ref{fig:examples}, it even generates a pedestrian with better shape. However, this model can not learn the adequate background context information to fit surrounding pixels.

We analyze the reason why the two discriminators $D_p$ and $D_b$ have different optimal GAN losses in our work: 1) for $D_p$, as we apply the $Patch$GAN trick, LSGAN with least square loss will get larger error than the regular GAN loss. It makes the model to be more sensitive to every pixel in images than the regular GAN. Thus the generator $G$ may be forced to learn too much detailed information of pedestrians instead of capturing the global distribution; 2) however, our discriminator $D_b$ can take benefit from the least square loss when learning the background context information. We expect the generator to strictly learn the background information from the surrounding pixels.

\begin{figure}
\centering
\subfigure[Generated pedestrians]{
\begin{minipage}{3.8cm}
\centering
\includegraphics[width=3.6cm, height=1.6cm]{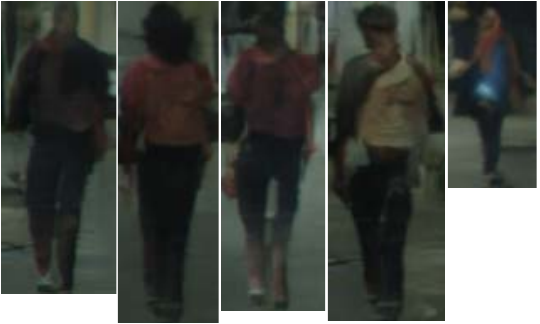}
\end{minipage}
}
\subfigure[Real pedestrians]{
\begin{minipage}{3.8cm}
\centering
\includegraphics[width=3.6cm, height=1.6cm]{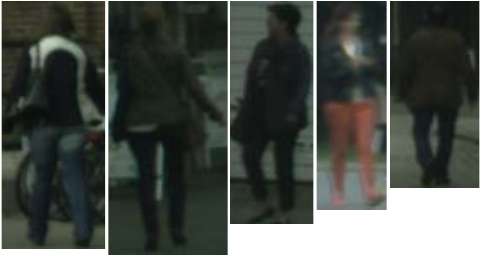}
\end{minipage}
}
 \caption{Comparison of generated and real pedestrians.}
 \label{fig:person}
\end{figure}

\begin{figure}[!t]
\centering
\includegraphics[width=8.5cm, height=4.4cm]{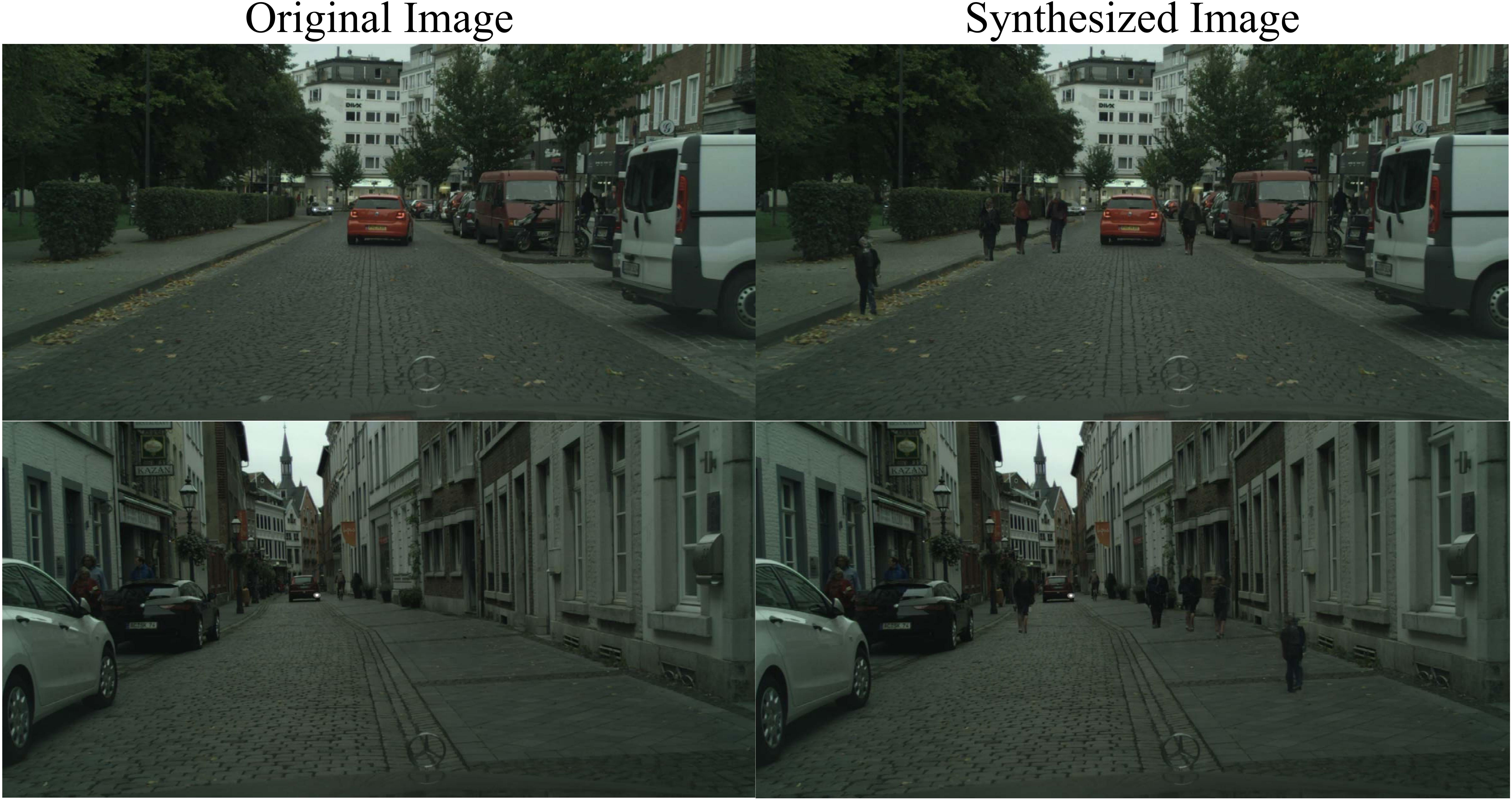}
\caption{The examples of synthesizing pedestrians in the original scenes. It shows the original images in the left and corresponding synthesized images in the right.}
\label{fig:big}
\vspace{-.8em}
\end{figure}

\begin{table}[h]
\footnotesize
\newcommand{\tabincell}[2]{\begin{tabular}{@{}#1@{}}#2\end{tabular}}
\caption{The performance comparison of using different settings to train the Faster R-CNN, including adding different amounts of synthetic data from $Pix2Pix$ GAN and PS-GAN, separately.} \label{compare_cityscapes}
\centering
\begin{tabular}{c|c|c}
\hline
\textbf{Data} & \tabincell{c}{\textbf{$\mathbf{Pix2Pix}$ GAN}\\ }  & \tabincell{c}{\textbf{PS-GAN}\\ } \\
\hline
\hline
\tabincell{c}{1826 real images (7729 labels)} & \multicolumn{2}{c}{60.11\%}\\
\hline
\tabincell{c}{+ 3000 synthetic pedestrians}& 59.95\%& 61.02\% \\
\tabincell{c}{+ 5000 synthetic pedestrians}& 60.23\%& \textbf{61.79\%} \\
\tabincell{c}{+ 8000 synthetic pedestrians}& 58.41\%& 61.59\% \\
\hline
\hline
Pascal VOC 2007 & \multicolumn{2}{c}{34.13\%}\\
Pascal VOC 2007 \& 2012 & \multicolumn{2}{c}{36.85\%}\\
\hline
\hline
\tabincell{c}{300 real images (1173 labels)} &\multicolumn{2}{c}{47.08\%} \\
\hline
\tabincell{c}{+ 500 synthetic pedestrians}& 46.97\% & 47.36\%\\
\tabincell{c}{+ 1000 synthetic pedestrians}& 46.71\%  & \textbf{48.79\%} \\
\tabincell{c}{+ 2000 synthetic pedestrians}& 46.12\% & 48.11\% \\
\hline
\hline
\tabincell{c}{1000 real images (4368 labels)} & \multicolumn{2}{c}{52.72\%}\\
\hline
\tabincell{c}{+ 2000 synthetic pedestrians}& 52.07\%& 54.41\% \\
\tabincell{c}{+ 4000 synthetic pedestrians}& 51.68\%& \textbf{56.19\%} \\
\tabincell{c}{+ 5000 synthetic pedestrians}& 51.24\%& 55.96\% \\
\hline
\end{tabular}
\end{table}

We crop the pedestrians from the generated images, and demonstrate that PS-GAN can generate pedestrians with sharp body shapes and detailed information as illustrated in Figure \ref{fig:person}. Compared with the work in \cite{zheng2017unlabeled}, which uses 12,936 images to train the GAN for the person re-identification task, we only use 1200 images to train PS-GAN and get sharper and more photo-realistic results.



\subsubsection{Quantitative Analysis} \label{cityscapes quantitative}
In this section, we combine the data generated by PS-GAN with some real data to train the Faster R-CNN detector \cite{faster_rcnn} to analyze the effects of data augmentation. In the experiment, we follow the setting on the above qualitative result section and randomly put noise boxes to generate pedestrians on the $256 \times 256$ patches from the images on Cityscapes. After that, we fill those patches with generated pedestrians into the original images. Some examples are shown in Figure \ref{fig:big}.
Many synthetic pedestrians by PS-GAN look hallucinating real in the original images, which is only trained on the 1826 training images. It is notable that all the patches are add into the original 1826 training images which means we do not involve any new images with synthetic pedestrians. To demonstrate how the augmented synthetic images can help boost the performance of the Faster R-CNN model, we train three Faster R-CNN detectors\cite{faster_rcnn} (VGG-16 \cite{simonyan2014very} based models). The baseline detector is trained on the original 1826 training images, and two detectors are trained on those images adding synthetic pedestrians from $Pix2Pix$ GAN and PS-GAN separately.
All the detectors are tested on the 500 testing images and the average precisions (AP) are from the best performance when all the models converge. We also add different amounts of synthetic pedestrians into the 1826 training image and present the results on Table \ref{compare_cityscapes}.
Although the Faster RCNN detector has been trained well (60.11\%) on 1856 images, adding synthetic pedestrians on the original images for training the detector is still beneficial.
With 5000 synthetic pedestrians from PS-GAN, we improve the the detector performance from 60.11\% to 61.79\%.
On the contrary, adding 8000 synthetic pedestrians from $Pix2Pix$ GAN downgrades the performance to be 58.41\% since adding too many examples from $Pix2Pix$ GAN destroys the normal data distribution. This experimental result matches the terrible visual quality of the $Pix2Pix$ GAN.



To attain deeper insight into the effect of the augmented synthetic images, we conduct more experiments as shown in Table \ref{compare_cityscapes}. We train a baseline Faster R-CNN detectors\cite{faster_rcnn} (VGG-16 \cite{simonyan2014very} based models) of using 300 real image and also adopt the detectors \cite{faster_rcnn} pretrained on Pascal VOC \cite{pascal-voc-2007}. Also, all the detectors are tested on the 500 testing images. Moreover, to avoid the GAN model to see more data than the Faster R-CNN, all the Pix2Pix GAN model and PS-GAN models are retrained on the same image set for training the Faster R-CNN. In other words, we retrain those GAN models on the 300 images for fair comparison. The synthetic pedestrians are also adding into the original images without adding any new image into training.
As shown in Table \ref{compare_cityscapes}, the detectors pretrained on Pascal VOC 2007 dataset and 2007 \& 2012 datasets can achieve 34.13\% AP and 36.85\% respectively. This observation indicates that pretrained model on different background can not perform well. The baseline detector using 300 real images with 1173 pedestrians in Cityscapes, can achieve 47.08\% of the average precision (AP) for pedestrian detection. By adding the synthetic images, the AP rate can be improved.
We get the best performance when adding 1000 synthetic pedestrians. It outperforms the baseline to 1.71\% while adding 2000 synthetic pedestrians can improve by only 1.04\%.
In both cases, we compare the results with image synthesized from $Pix2Pix$ GAN. It slightly downgrades the performance in all the experiments.

We also train another baseline Faster R-CNN detector using 1000 real images, with 4368 pedestrians annotated in total. Meanwhile, the GAN models are retrained on the 1000 real images. The motivation here is to see how the augmented synthetic images can help boost the performance when the Faster R-CNN model gets different amounts of real training data.
We add 2000 and 4000 synthetic pedestrians to the original 1000 real ones and retrain the Faster R-CNN detector.
We can see that, even Faster R-CNN is trained in a more saturated state, the model with data augmentation can achieves 56.19\% AP, outperforming the baseline 3.47\%.

\subsection{Tsinghua-Daimler Cyclist Benchmark}
Tsinghua-Daimler Cyclist Benchmark \cite{Tsinghua} is a dataset for cyclist detection, which contains 4 subsets: train, validation, test and ``NonVRU" set. The train set contains 9741 images with annotations as ``cyclist".  There are 1019 images in validation set and 2914 images in test set, which contain the annotations as ``pedestrian", ``cyclist", ``motorcyclist", ``tricyclist", ``wheelchairuser", and ``mopedrider". The ``NonVRU" set contains 1000 images with background image only (no pedestrian).



To explore the generalization ability of PS-GAN, we perform the cross-dataset test. The goal of this experiment is to simulate the situation that applying our GAN model on the new unannotated video or video with limited supervision. It is useful to improve the performance when the training set contains the similar scenes in testing set. If the PS-GAN has great generalization ability in the new data, it could be very helpful when we face a new task with limited annotated information.

\begin{figure}
\centering
\subfigure[Synthesize pedestrians in background images.]{
\begin{minipage}{5.8cm}
\includegraphics[width=5.8cm, height=5.3cm]{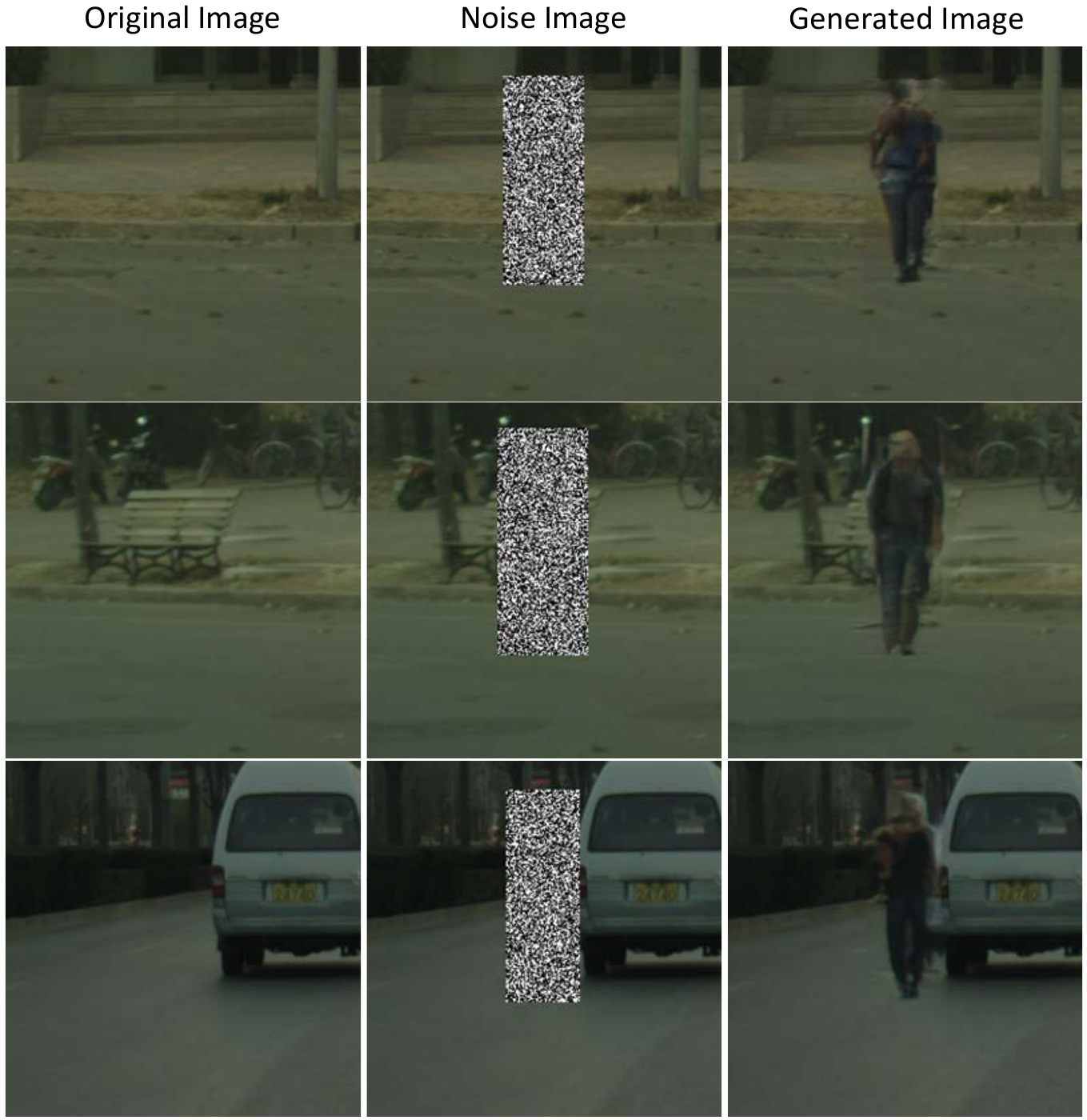}
\label{fig:Tsinghua:a}
\end{minipage}
}
\subfigure[Synthetic images for data augmentation]{
\begin{minipage}{5.8cm}
\includegraphics[width=5.8cm, height=3.6cm]{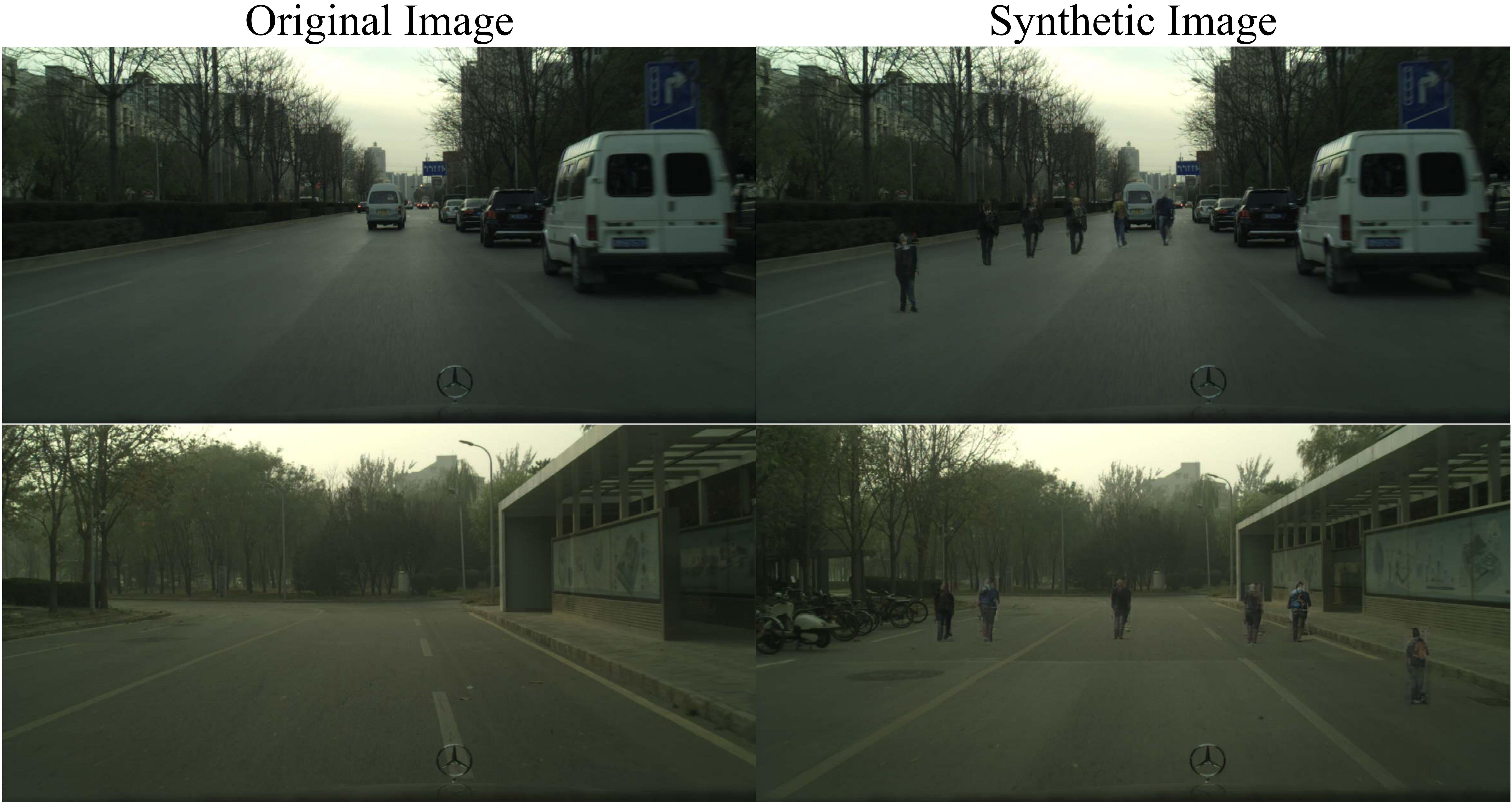}
\label{fig:Tsinghua:b}
\end{minipage}
}
 \caption{The results of generating pedestrians on Tsinghua-Daimler Cyclist Benchmark. All those images are generated by PS-GAN pretrained on Cityscapes without any data from Tsinghua-Daimler Cyclist Benchmark.}
 \label{fig:Tsinghua}
\end{figure}

Firstly, we directly apply the PS-GAN model pretrained on Cityscapes (using 1826 images) to generate pedestrians on the empty background images from ``NonVRU" set. Since some images in ``NonVRU" set are not suitable (e.g. no road, too dark or light, etc.) to synthesize pedestrians, we get 650 images after removing those images. Similar to what we did in Cityscapes, we cropped $256 \times 256$ patches from those images and put noise boxes to synthesize pedestrians. The generated examples are shown in Figure \ref{fig:Tsinghua:a}. \textbf{Without adding any data from Tsinghua-Daimler Cyclist Benchmark, PS-GAN can still generate high-quality and realistic images on this dataset.} Note that there are many differences between these two datasets, such as the background, lighting conditions and pedestrian styles. We can expect the generated image quality has a slight drop compared to the results in Cityscapes. Specifically, the region around pedestrian does not match the background well, and the body of pedestrian loses some details in some cases. Nevertheless,
the generated images still look natural with satisfactory qualities.


Also, we conduct the comparison between using the real data on Cityscapes and adding synthetic data to train the Faster R-CNN. For the test, all the 2914 test images of Tsinghua-Daimler Cyclist Benchmark with the bounding boxes annotated as ``pedestrian" and ``cyclist" are directly used. The results are presented in Table \ref{compare_Tsinghua}, where adding the 650 synthetic images gain a huge improvement (2.64\%) than the baseline with the real data on Cityscapes. Different from the setting in Cityscapes, we add new background images when adding the synthetic pedestrians. To illustrate the effect of adding new images, we also compare with the detector trained on the real data on Cityscapes and the 650 empty background images. Adding background images can bring a slight improvement, about 0.29\%. In this case, the result with image synthesized from $Pix2Pix$ GAN can slight improve the AP rate but the improvement is much poor compared with the PS-GAN by 2.3\%.

Meanwhile, we execute the detection experiments with different amounts of training data for the Faster-RCNN. We report the results of using 300 and 1000 real images, and also adding synthetic images and background images separately on Table \ref{compare_Tsinghua}. Also, we use the GAN models retrained on 300 and 1000 images as we did in section \ref{cityscapes quantitative}. The performances get improved in both cases. Adding background images can bring limited improvement, 0.91\% and 0.6\%, respectively. Adding some synthetic data here shows significant help here, boosting the performance by 2.62\% and 2.52\%, respectively. Especially when adding 650 synthetic images into 1000 real images, the AP rate get better from 42.42\% to 44.94\% which even significantly outperforms the AP rate 43.77\% of using 1826 real images to train the detector. Moreover, in all cases, we compare the results with image synthesized from $Pix2Pix$ GAN. It can only achieve similar AP rate as the baseline detectors and has not done better than PS-GAN.

\begin{table}[!t]
    \small
    \newcommand{\tabincell}[2]{\begin{tabular}{@{}#1@{}}#2\end{tabular}}
    \caption{The comparison of adding different amounts of synthetic data to train the Faster R-CNN.
    The number of label means the number of real pedestrians in the real images or generated pedestrians in the synthetic images.
}
    \label{compare_Tsinghua}
    \centering
    \begin{tabular}{c|c|c}
\hline
\textbf{Data} & \tabincell{c}{\textbf{$\mathbf{Pix2Pix}$ GAN}\\ }  & \tabincell{c}{\textbf{PS-GAN}\\ } \\
        \hline
        \hline
        1826 real images from Cityscapes  $^\star$ (7729 labels)  & \multicolumn{2}{c}{43.77\%} \\
        \hline
        + 650 background images (no pedestrian) & \multicolumn{2}{c}{44.06\%}\\
        \hline
        + 650 synthetic images (4500 pedestrians) & 44.11\% & \textbf{46.41\%}\\
        \hline
        \hline
        Pascal VOC 2007 & \multicolumn{2}{c}{23.24\%} \\
        Pascal VOC 2007 \& 2012 & \multicolumn{2}{c}{26.50\%} \\
        \hline
        \hline
        300 real images from Cityscapes $^\star$ (1173 labels)  & \multicolumn{2}{c}{32.15\%} \\
        \hline

     + 300 background images (no pedestrian) & \multicolumn{2}{c}{33.06\%} \\
     \hline

     + 300 synthetic images (2000 pedestrians) & 32.64\% & \textbf{34.77\%}\\
        \hline
        \hline
        1000 real images from Cityscapes  $^\star$ (4368 labels)  & \multicolumn{2}{c}{42.42\%} \\
        \hline
        + 650 background images (no pedestrian)  & \multicolumn{2}{c}{43.02\%} \\
        \hline
        + 650 synthetic images (4500 pedestrians) & 42.70\% & \textbf{44.94\%} \\
        \hline
    \end{tabular}
\end{table}

\begin{table}[h]
\footnotesize \caption{The AP rate comparison of different pretrained detectors. The two sets, Cityscapes and Tsinghua-Daimler Cyclist, are used as the background.} \label{tab:ap:cmp}
\centering
\begin{tabular}{ccccc}
\toprule
\multirow{2}{*}{\textbf{Generator}} & \textbf{Pretrained} & \multicolumn{2}{c}{\textbf{Background}}  \\
\cline{3-4}
& \textbf{Detector} & Cityscapes & Tsinghua \\
\midrule
PS-GAN & Pascal VOC & 84.55\% & 88.85\% \\
& Cityscapes & 90.11\% & 90.46\% \\
\midrule
$Pix2Pix$ GAN& Pascal VOC & 52.46\% & 69.42\% \\
& Cityscapes & 58.82\% & 71.68\% \\
\bottomrule
\end{tabular}
\end{table}

\subsection{Evaluation with Pretrained Detectors}
Finally, we use the detectors pretrained on real images to detect the synthetic samples (using 500 samples) and report the AP rate. Two Faster RCNN detectors \cite{faster_rcnn} trained on Pascal VOC and Cityscapes (300 samples) are utilized. We also compare PS-GAN with $Pix2Pix$ GAN on this task. The results are list in Table \ref{tab:ap:cmp}. We can see that the AP rate of the detectors on the samples generated with PS-GAN are much higher than that with $Pix2Pix$ GAN, showing the generation power of PS-GAN in another prospective.

\section{Conclusion}
We propose PS-GAN to synthesize pedestrian within the certain bounding boxes in real scenes. The experimental results show that our model can generate high quality pedestrian images, and the synthetic images can effectively improve the ability of the CNN based detectors. In the cross dataset test, our PS-GAN model trained on Cityscapes can do pretty good generation in the other new dataset as well as help boost the detection, which demonstrates the ability of generalization and transferring knowledge. This is helpful when we face a new task with limited annotated information.

Currently PS-GAN pedestrians vary in a mild range of scales (can not be too small or large), which restricts it to generate more diverse and natural data. Making PS-GAN to handle the extreme case is challenging. Besides that, how to control PS-GAN to generate pedestrians in reasonable locations (e.g., pedestrian should not be on the tree or in the water) is also interesting.

In the meantime, applying PS-GAN to other detection tasks is definitely one of our future works.

\clearpage

\bibliographystyle{splncs}
\bibliography{egbib}

\section{Supplementary Experiments}
In this supplemental, We provide more results to help
understand our proposed approach described in the paper.
Firstly, we show more generation results from PS-GAN on Cityscapges and Tsinghua-Daimler Cyclist Benchmark. Secondly, we investigate the effects of data augmentation used to boost the detection performance on both dataset.

\subsection{Generated images comparison}
\begin{figure*}[!h]
\centering
\includegraphics[width=12cm, height=5cm]{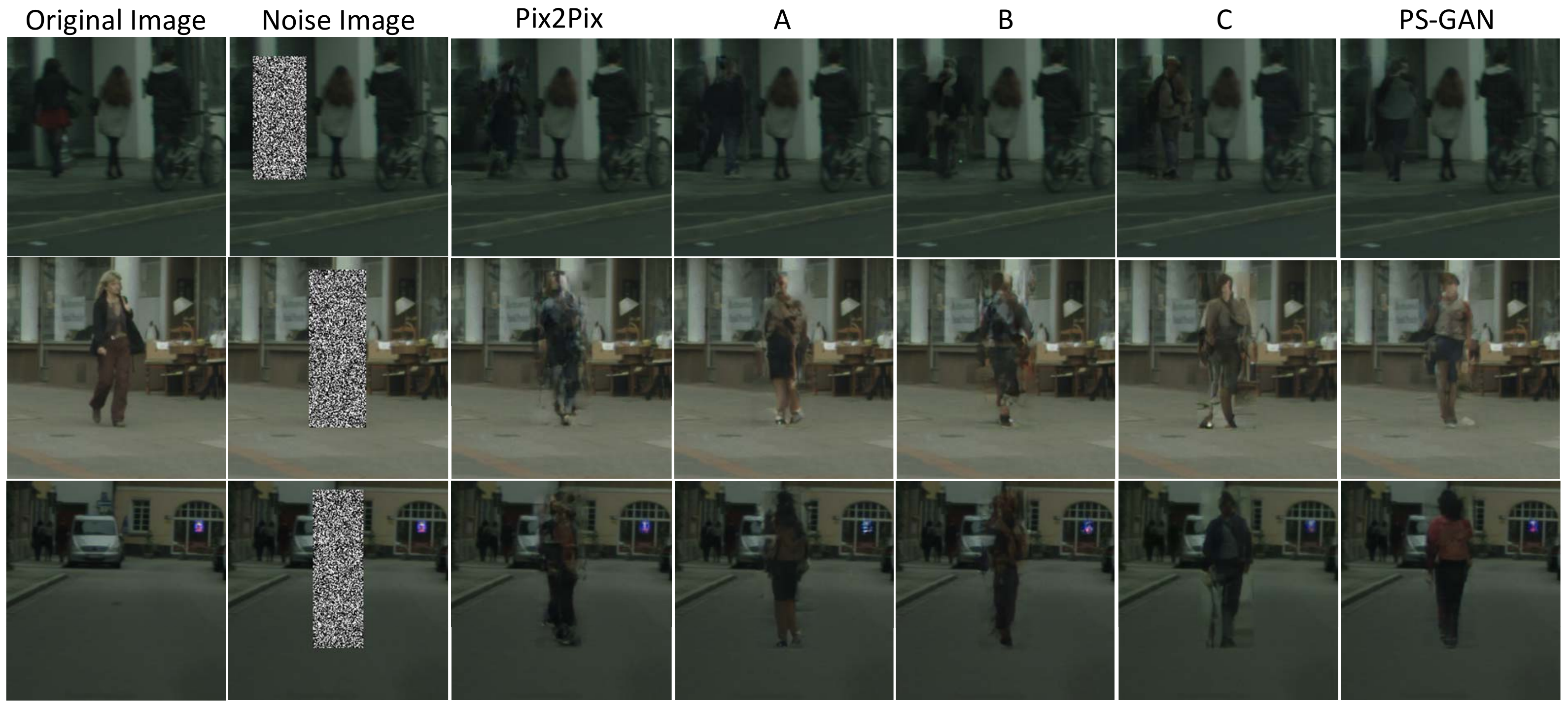}
\caption{More results of synthesizing pedestrians on cityscapes with different models.}
\label{fig:gen:compare_cityscapes}
\vspace{-.8em}
\end{figure*}

We show more results of the work in Section 4.1.1 in Figure \ref{fig:gen:compare_cityscapes}. PS-GAN still gets the best performance.
We also compare the synthetic images of all models on Tsinghua-Daimler Cyclist. Here we further evaluated different approaches in the cross-dataset setting.

\begin{figure*}[!htp]
\centering
\includegraphics[width=12cm, height=9cm]{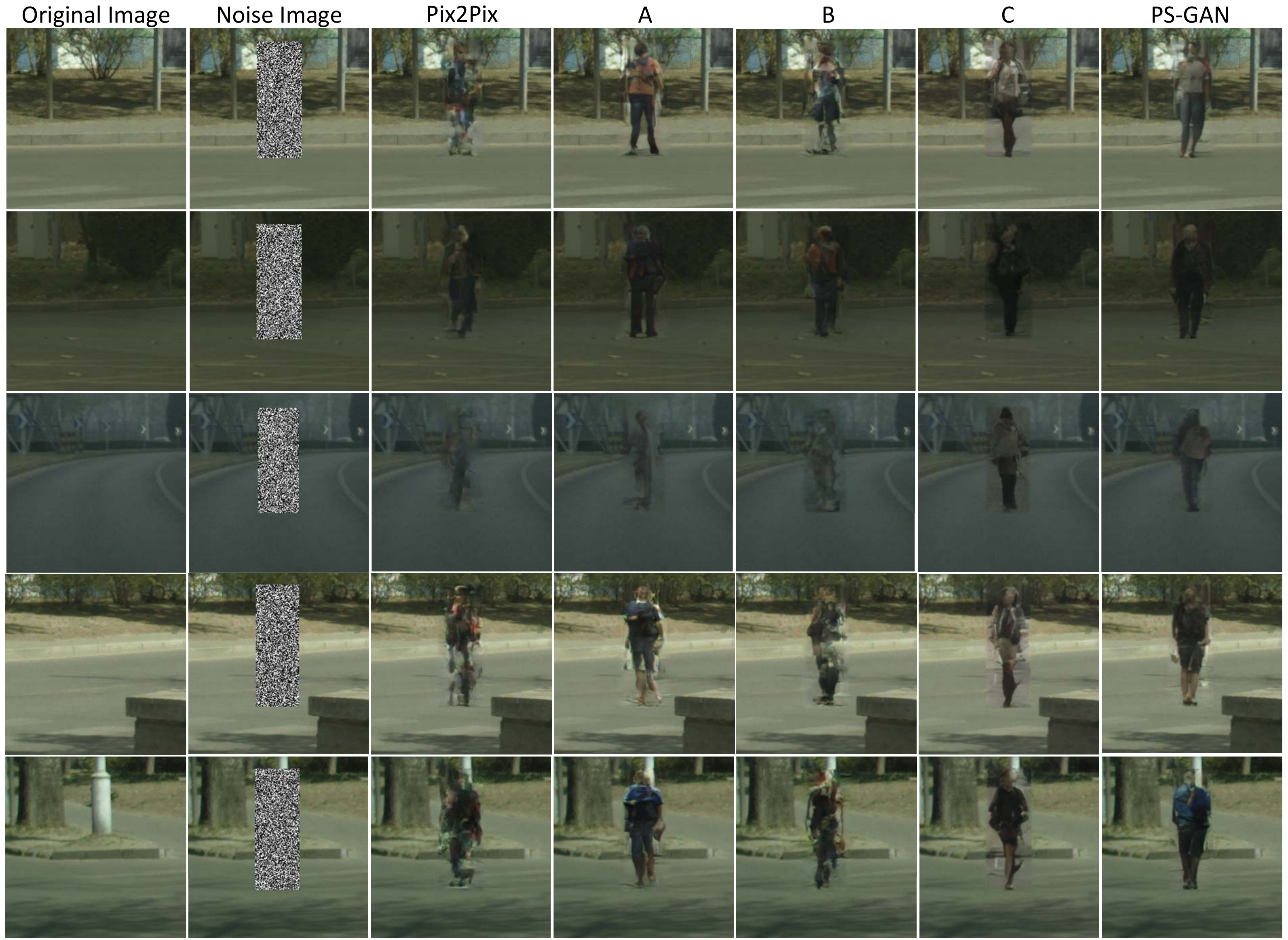}
\caption{Results of synthesizing pedestrians on Tsinghua-Daimler Cyclist Benchmark with different models.}
\label{fig:gen:compare_Tsinghua}
\vspace{-.8em}
\end{figure*}

As shown in Figure \ref{fig:gen:compare_Tsinghua}, PS-GAN, even trained on Cityscapes, can generate the best quality of images and fit the background well. on the contrary, the synthesized pedestrians using $Pix2Pix$ GAN are not good. Some of them are very blur. We also see the results of other variant ions of PS-GAN and have similar observation of them as in Cityscapes.

\subsection{Visualization of detection results}
To future investigate how the synthesized can help improving the performance of Faster R-CNN, we put some visualized detection results on for bot Cityscapes and Tsinghua-Daimler Cyclist. The experimental details are described in Section 4.1.2 and 4.2.

\begin{figure*}
\centering
\includegraphics[width=12cm, height=13cm]{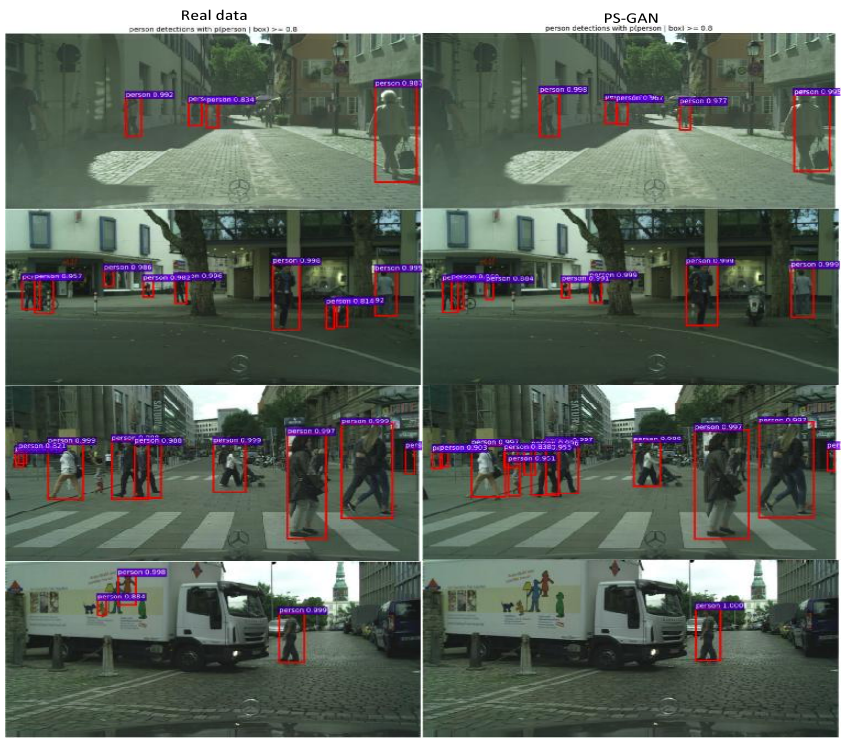}
\caption{Visualization of the detection result on Cityscapes. The left column contains the results of the Faster R-CNN trained on 300 real images, while the right are the results for adding 1000 synthetic pedestrians from PS-GAN.}
\label{fig:cityscapes:300}
\vspace{-.8em}
\end{figure*}

\begin{figure*}
\centering
\includegraphics[width=12cm, height=13cm]{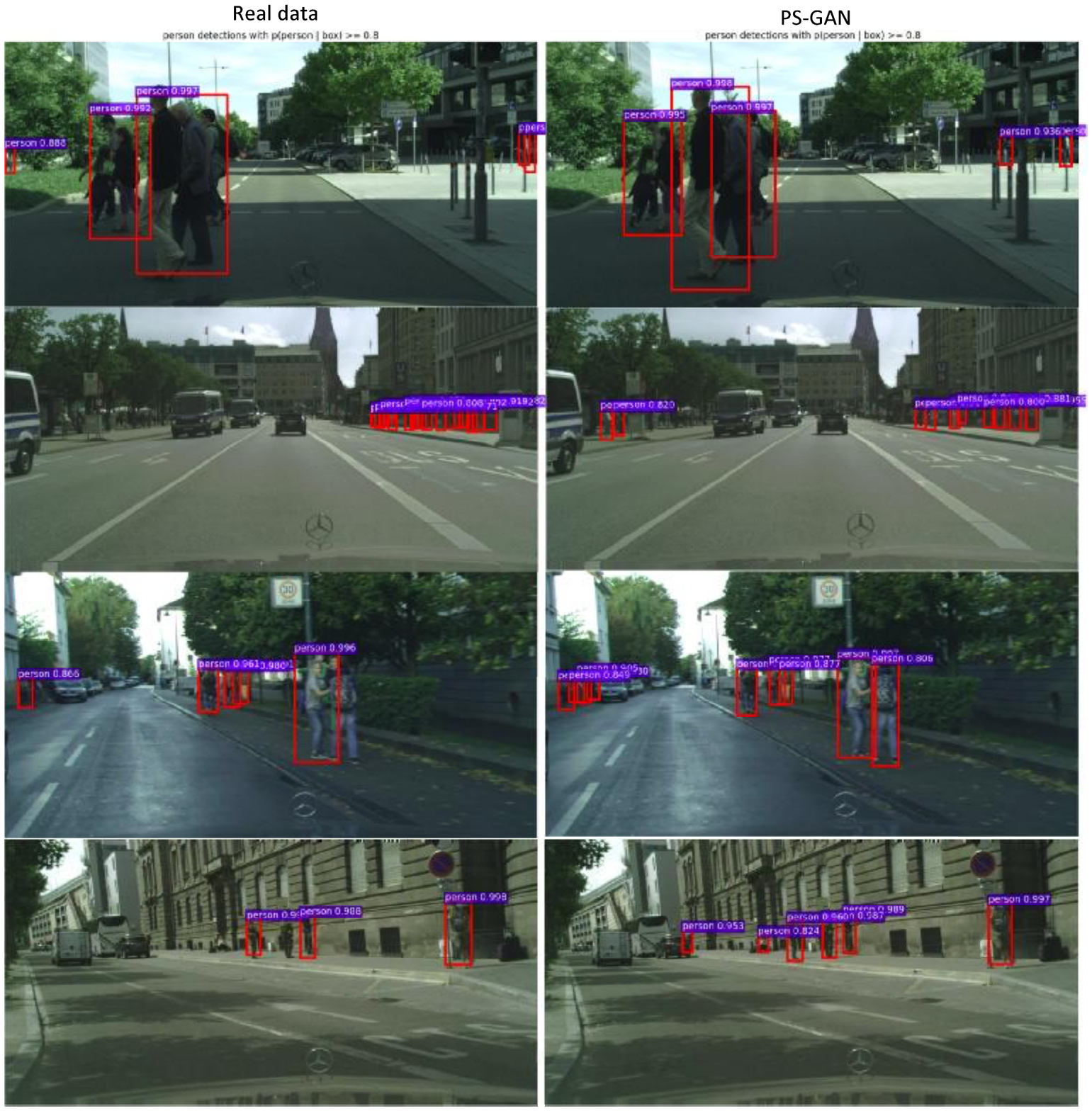}
\caption{Visualization of the detection result on Cityscapes. The left column contains the results of the Faster R-CNN trained on 1000 real images, while the right are the results for adding 4000 synthetic pedestrians from PS-GAN.}
\label{fig:cityscapes:1000}
\vspace{-.8em}
\end{figure*}

We first show some examples of Cityscapes in Figures \ref{fig:cityscapes:300} and  \ref{fig:cityscapes:1000}. For the 300 real image setting, it is clear that the data augmentation step can increase true positives while reduce some false positives. Even for the 1000 real setting, it can also gain more real detections than the original model.

\begin{figure*}
\centering
\includegraphics[width=12cm, height=13cm]{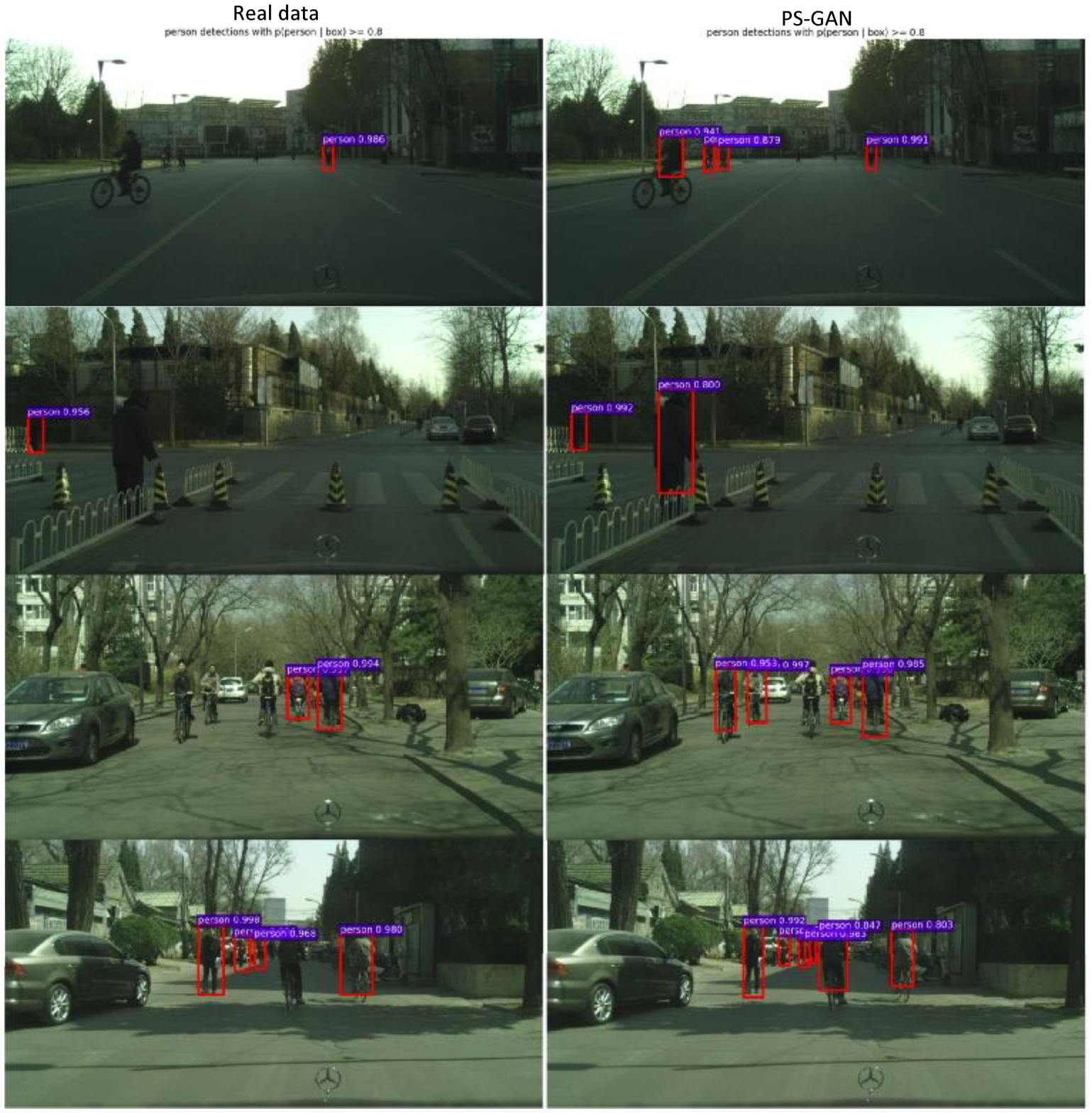}
\caption{Visualization of the detection result on Tsinghua-Daimler Cyclist Benchmark. The left column contains the results of the Faster R-CNN trained on 300 real images from Cityscapes, while the right are the results for adding 300 synthetic images from PS-GAN.}
\label{fig:compare_Tsinghua:300}
\vspace{-.8em}
\end{figure*}

\begin{figure*}
\centering
\includegraphics[width=12cm, height=13cm]{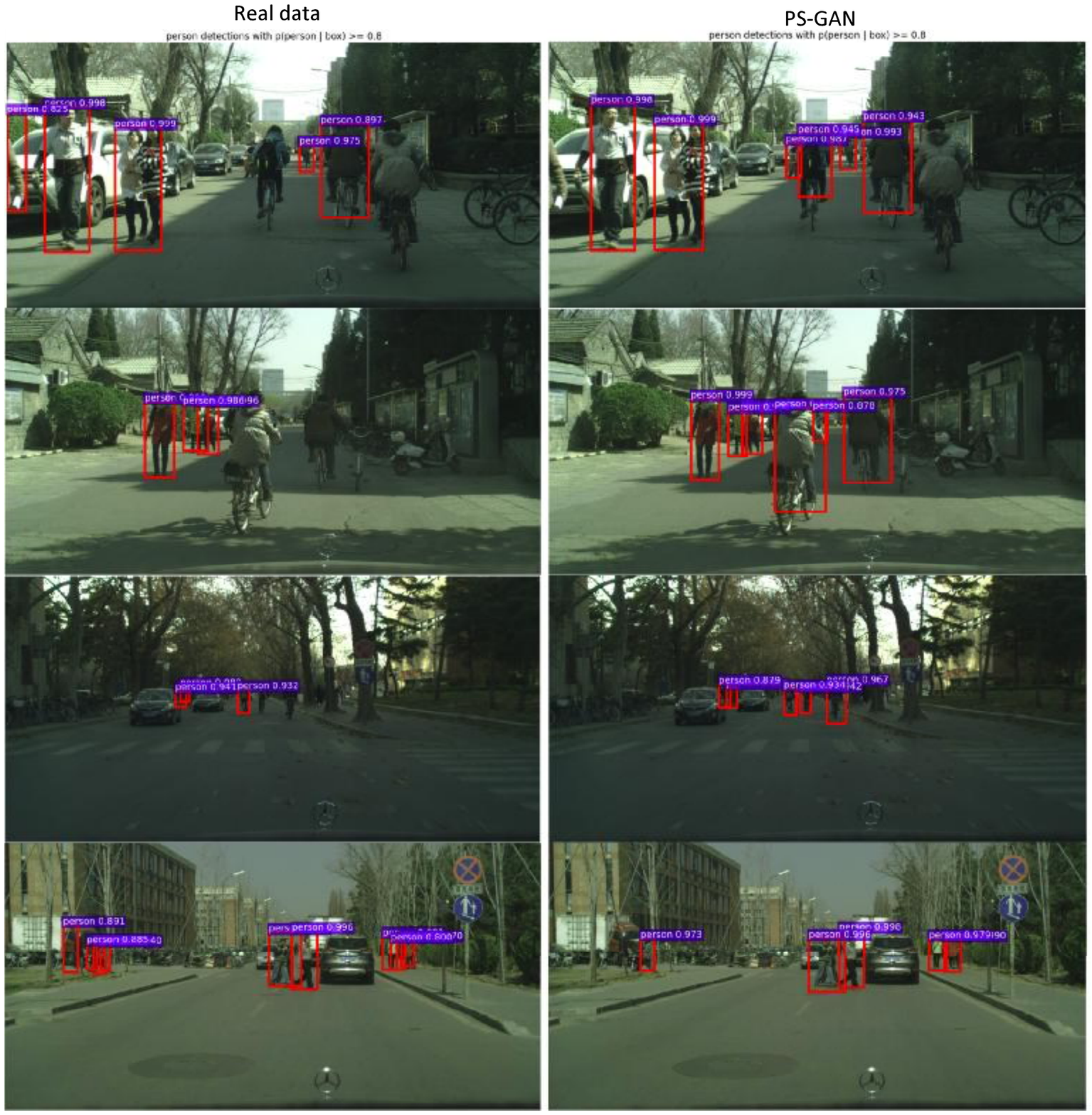}
\caption{Visualization of the detection result on Tsinghua-Daimler Cyclist Benchmark. The left column contains the results of the Faster R-CNN trained on 1000 real images from Cityscapes, while the right are the results for adding 650 synthetic images from PS-GAN.}
\label{fig:compare_Tsinghua:1000}
\vspace{-.8em}
\end{figure*}

For Tsinghua-Daimler Cyclist, the results are demonstrated in Figures \ref{fig:compare_Tsinghua:300} and  \ref{fig:compare_Tsinghua:1000}. In both settings, adding the generated data is able to boost the performance of detector by getting more real detections.

\end{document}